\RequirePackage[svgnames,table]{xcolor}
\documentclass[11pt,letterpaper]{yalearxiv}

\usepackage{microtype}
\usepackage{natbib}
\setcitestyle{authoryear,round}
\usepackage{array}
\usepackage{longtable}
\usepackage{pdflscape}
\usepackage{tikz}
\usepackage{multirow}

\hypersetup{
  colorlinks=true,
  linkcolor=blue!50!black,
  citecolor=blue!50!black,
  urlcolor=blue!50!black,
  pdftitle={Beyond Geometric Complementarity: Coherent Overlap in Sparse Mixture-of-Experts Routing},
  pdfauthor={Huiyuan Tian, Bonan Xu, Shijian Li},
  pdfsubject={Sparse mixture-of-experts routing and geometric complementarity},
  pdfkeywords={mixture-of-experts, sparse routing, expert subspaces, geometric complementarity}
}

\fancypagestyle{landscapetable}{%
  \fancyhf{}%
  \fancyfoot[C]{%
    \begin{tikzpicture}[remember picture,overlay]
      \node[rotate=90] at ([xshift=-0.50in]current page.east)
        {\footerfont\thepage};
    \end{tikzpicture}%
  }%
}

\title{Beyond Geometric Complementarity:\\Coherent Overlap in Sparse Mixture-of-Experts Routing}
\runningtitle{Coherent Overlap in Sparse Mixture-of-Experts Routing}

\author{%
Huiyuan Tian\textsuperscript{1},
Bonan Xu\textsuperscript{2}, and
Shijian Li\textsuperscript{1,*}\\
\textsuperscript{1}College of Computer Science and Technology, Zhejiang University, Hangzhou, China\\
\textsuperscript{2}Department of Aeronautical and Aviation Engineering, The Hong Kong Polytechnic University\\
\href{mailto:tianhuiyuan@zju.edu.cn}{\texttt{tianhuiyuan@zju.edu.cn}}\quad
\href{mailto:bonan.xu@polyu.edu.hk}{\texttt{bonan.xu@polyu.edu.hk}}\quad
\href{mailto:shijianli@zju.edu.cn}{\texttt{shijianli@zju.edu.cn}}\\
\textsuperscript{*}Corresponding author
}

\begin{document}

% yalearxiv.cls captures the abstract before \maketitle.
\begin{abstract}
\textbf{Abstract. }Sparse mixture-of-experts (MoE) language models route each token to multiple experts, suggesting a geometric account of their benefit: co-selected experts should contribute distinct representation directions. Existing evidence often conflates route coherence, candidate quality, and candidate-by-context interaction. We distinguish these quantities using an Expert Subspace Separation Index (ESSI), matched-route residuals, and a prefix-controlled \(2\times2\) factorial; frozen-route interventions and a controlled Top-\(k\) study assess functional value.
Three paired contrasts organize the findings. First, across six MoE architectures, expert subspaces overlap substantially, yet actual routes explain token representations better than matched alternatives. Second, across the 39 factorial cells in OLMoE, Mixtral, and DeepSeek, the selected candidate explains more of the residual representation than the strongest unselected rival in every cell, yet the actual prefix narrows this advantage throughout: all interactions are negative, and every 95\% confidence interval lies below zero. Third, this geometric narrowing does not imply functional redundancy: adding later experts improves next-token prediction in 24 of 39 frozen-route comparisons, while the other 15 estimates are inconclusive; a controlled training study also favors Top-2 over Top-1 in all three seeds.
We call this joint pattern \textbf{coherent overlap}: routing selects token-relevant experts from a shared geometric neighborhood, while useful multi-expert computation persists without disjoint linear coverage. Separating these quantities clarifies why geometric similarity alone cannot determine redundancy or pruning value.

\end{abstract}

\maketitle

%% ================================================================
\section{Introduction}\label{sec:intro}

Mixture-of-experts (MoE) models combine input-dependent gating with
specialized local computations~\citep{jacobs1991adaptive,jordan1994hierarchical}.
Sparse variants scale total capacity while activating only a small
parameter fraction per token~\citep{shazeer2017outrageously,
lepikhin2020gshard,du2022glam,fedus2021switch,zoph2022stmoe}. Routing
spans learned token choice, balanced assignment, deterministic hashing,
and expert choice~\citep{fedus2021switch,lewis2021base,roller2021hash,
zhou2022expertchoice}; scaling analyses distinguish total capacity,
active compute, and expert granularity~\citep{clark2022scaling,
ludziejewski2024finegrained}. Modern language models commonly use
learned Top-\(k\) routing~\citep{jiang2024mixtral,muennighoff2024olmoe,
dai2024deepseekmoe}. Their success makes a basic mechanistic question
consequential: \textbf{what does a token gain from being processed by
several experts rather than one?}

\begin{figure}[t]
    \centering
    \includegraphics[width=\columnwidth]{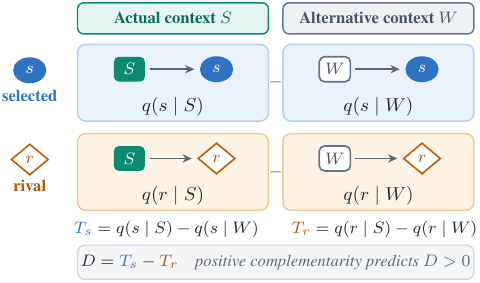}
    \caption{The \(2\times2\) design isolates the
        candidate-by-context interaction~\(D\).}
    \label{fig:method}
\end{figure}

A common answer is \emph{geometric complementarity}.
Specialization-oriented MoE designs aim to distribute focused,
non-overlapping knowledge across experts, while analyses of pretrained
MoEs report diverse behavior, structured co-activation, router--expert
geometric coupling, and routing-induced representation collapse~\citep{
dai2024deepseekmoe,lo2024closer,tang2025collaboration,ahrac2026geometry,
chi2022collapse}. These observations suggest a simple mechanism: one selected expert captures the token's
dominant directions and the remaining experts cover what is left. The
intuition offers both an explanation for
multi-expert quality and a design principle for sparse computation. It also motivates expert pruning and compression~\citep{lu2024notall,
zhang2025diversifying,li2026submoe,hu2026mosaic} and objectives for diverse
or stable specialization~\citep{do2025simsmoe,liu2024orthogonal,
park2026mass}.

The conditional claim behind that intuition is stronger than the
observations usually offered in its support. In particular, it combines
three properties that need not coincide:
\begin{enumerate}
\item \textbf{Route coherence}: the complete selected set fits the
  token better than a matched alternative set.
\item \textbf{Candidate quality}: under a fixed context, the selected
  expert fits the token better than the strongest unselected alternative.
\item \textbf{Positive geometric complementarity}: the selected expert
  becomes \emph{especially} useful because of the specific experts with
  which it is grouped.
\end{enumerate}
A coherent route may simply contain individually strong candidates.
Likewise, a selected candidate may be strong even when the preceding
experts have already covered most of the directions it can add.

Prior work studies knowledge attribution, linguistic and semantic routing,
diversity, co-activation, router--expert geometry, and route
counterfactuals~\citep{wang2026deconstructing,antoine2025pos,
olson2025semantic,lo2024closer,tang2025collaboration,wang2026myth,
wang2026illusion,ahrac2026geometry,yoon2026misrouted}.
These lines of work characterize internal structure or route quality, but
leave unresolved whether a candidate expert's advantage depends on its
co-selection context. Identifying the
interaction requires crossing candidates and contexts while holding
each fixed in turn (Figure~\ref{fig:method}).

We build that identification in two stages. First, ESSI calibrates
between-expert distance by local within-expert tangent dispersion,
because raw subspace distance has no natural baseline. Second, a prefix-controlled \(2\times2\) factorial crosses the
selected candidate and strongest unselected rival with the actual and a
matched alternative prefix. Its difference-in-differences tests the
defining prediction of positive geometric complementarity: the selected
candidate's advantage should be larger in its actual co-selection
context.

We apply the factorial and functional protocols to OLMoE
(routed-only, Top-8/64), Mixtral (routed-only, Top-2/8), and
DeepSeek-MoE (shared-expert, Top-6/64)~\citep{muennighoff2024olmoe,
jiang2024mixtral,dai2024deepseekmoe}; the geometry survey also includes
Qwen3, Gemma4, and Qwen3.6~\citep{qwen3report,gemma4report,qwen36release}.
Across these models, expert subspaces overlap while actual routes fit
tokens better than matched alternatives (Section~\ref{sec:overlap}). In
the factorial analyses, selected candidates explain more of the residual
than their rivals, but actual prefixes narrow rather than amplify that advantage
(Section~\ref{sec:complementarity}). Direct interventions show that later
experts can remain functionally useful despite low marginal geometric
novelty (Section~\ref{sec:function}). We call this combination
\textbf{coherent overlap}: routing is structured and valuable, but its
value is not explained by assigning a disjoint linear piece of the
representation to each selected expert.

This work advances MoE routing analysis in three ways:
\begin{itemize}
\item \textbf{A diagnostic framework for geometric complementarity.}
  We introduce ESSI to normalize between-expert separation by local
  within-expert geometry, and a prefix-controlled \(2\times2\) factorial
  that separately identifies candidate quality, contextual opportunity,
  and candidate-by-context interaction. The framework tests
  complementarity rather than inferring it from a good route.
\item \textbf{A cross-architecture empirical pattern.}
  Across six architectures, expert subspaces overlap while actual routes
  remain coherent. Across the 39 factorial cells in OLMoE, Mixtral, and
  DeepSeek, selected candidates explain more of the residual than their
  rivals, but the actual context consistently narrows this advantage.
\item \textbf{A separation between geometric novelty and functional value.}
  Later experts reduce next-token NLL in 24 of 39 frozen-route additions;
  the remaining estimates are statistically inconclusive. A controlled
  Top-1/Top-2 training study also favors Top-2 in all three seeds. Together, these results show why output-level interventions are needed before using
  input-subspace overlap as a proxy for redundancy, pruning value, or
  multi-expert benefit.
\end{itemize}

\begin{figure*}[t]
	\centering
	\includegraphics[width=\textwidth]{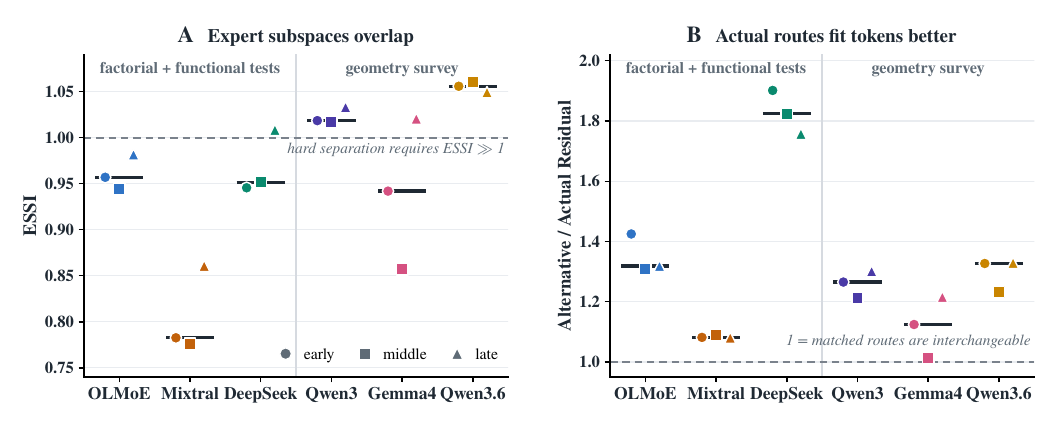}
	\caption{Experts overlap, but actual routes are coherent.
		A:~ESSI (Eq.~\ref{eq:essi}) remains near one across six
		models and three layers each (range 0.776--1.060, median 0.969),
		indicating that between-expert separation is comparable to local
		within-expert variation.
		B:~The residual ratio (alternative divided by actual) exceeds
		one in all 18~cells (range 1.012--1.901, median 1.283), showing
		that actual routes fit the token better than matched alternatives.
		Both panels include only experts meeting the support threshold.}
	\label{fig:overlap}
\end{figure*}

\section{Method}\label{sec:method}
%% ================================================================

\subsection{Measuring Geometric Fit}\label{sec:geometric-score}

At layer~\(\ell\), let \(z_x^\ell\in\mathbb{R}^d\) denote the
router-input representation for token~\(x\). The router selects an
ordered set of \(k\) experts,
\(E_x^\ell=(e_1,\ldots,e_k)\), in decreasing router-score order. We call \(e_1\) the \textbf{leader} and
\(e_{2:k}\) the \textbf{later experts}. The score-ordered prefix before
position~\(j\) is
\begin{equation}
S_{j-1}=(e_1,\ldots,e_{j-1}).
\end{equation}
The ordering is used only for attribution: in a standard Top-\(k\) MoE,
all selected experts execute in parallel. We suppress the layer index
below.

Global separation and context-conditioned marginal coverage answer
different questions, so we fit two subspace families. For ESSI, each eligible routed expert~\(e\) receives a
\emph{centered} rank-\(p\) global basis
\(G_e\in\mathbb{R}^{d\times p}\), fitted from train-split router inputs
with native Top-\(k\) weights renormalized within each token. Around each
sampled expert anchor~\(a\), we also fit a centered local tangent basis
\(L_{e,a}\) from its Euclidean-nearest assigned representations.
We use \(p=128\) throughout; anchor counts, neighborhood size, support
thresholds, and sampling rules are given in Appendix~\ref{app:essi}.

SVCCA, PWCCA, and CKA compare activation
spaces~\citep{raghu2017svcca,morcos2018pwcca,kornblith2019cka}; here the
objects are equal-rank linear spans, so principal-angle geometry is the
direct comparison.

For two rank-\(p\) bases \(U\) and \(V\), let
\(\theta_1,\ldots,\theta_p\) be their principal angles. We use the
normalized chordal (projection) distance on the Grassmann
manifold~\citep{edelman1998geometry,ye2016schubert},
\begin{equation}\label{eq:grassmann}
\begin{aligned}
d_{\mathrm{ch}}(U,V)
&=\left(\frac{1}{p}\sum_{i=1}^{p}\sin^2\theta_i\right)^{1/2}\\
&=\frac{\|UU^\top-VV^\top\|_F}{\sqrt{2p}}.
\end{aligned}
\end{equation}
Let \(\lambda_e\) be expert~\(e\)'s routing-load share among eligible
experts and \(\mathcal A_e\) its valid anchors. We introduce the
\textbf{Expert Subspace Separation Index} (ESSI),
\begin{equation}\label{eq:essi}
\mathrm{ESSI}=
\frac{\operatorname*{mean}_{e<f}d_{\mathrm{ch}}(G_e,G_f)}
{\max\!\left\{
 \sum_e\lambda_e\operatorname*{mean}_{a\in\mathcal A_e}
 d_{\mathrm{ch}}(G_e,L_{e,a}),\,\epsilon\right\}},
\end{equation}
with \(\epsilon=10^{-12}\). The numerator is unweighted across eligible
expert pairs; the denominator is load-weighted across experts. ESSI
therefore compares inter-expert separation with the local dispersion
already present inside each expert's routed region. An ESSI near one
means that inter-expert separation is comparable to this reference
variation; values well above one indicate stronger separation.

To measure candidate novelty conditional on a route prefix, the
factorial experiment instead uses a separate \emph{uncentered} rank-128
basis \(B_e\) for each eligible expert. For an expert set~\(P\),
let
\begin{equation}
B(P)=\operatorname{orth}\!\bigl([B_e:e\in P]\bigr).
\end{equation}
For nonzero-energy router input~\(z_x\), the unexplained energy fraction
is
\begin{equation}\label{eq:rho}
\rho_x(P)=
\frac{\|(I-B(P)B(P)^\top)z_x\|_2^2}{\|z_x\|_2^2}.
\end{equation}
The denominator is the raw input energy used by the implementation;
rows with zero input energy are excluded.
Adding candidate~\(c\) yields the \textbf{fractional
novelty}
\begin{equation}\label{eq:q}
q_x(c\mid P)=
\frac{\rho_x(P)-\rho_x(P\cup\{c\})}
     {\max\!\bigl(\rho_x(P),\epsilon\bigr)},
\qquad \epsilon=10^{-12}.
\end{equation}
If \(q=0.2\), the candidate explains 20\% of what remained after~\(P\).
A candidate aligned with the token can still have low~\(q\) when its
relevant directions are already covered by the context. Appendix~\ref{app:factorial-protocol} details the basis fit,
eligibility, masking, and numerical checks.

\begin{table*}[t]
	\centering
	\small
	\begin{tabularx}{\textwidth}{@{}
			>{\raggedright\arraybackslash}X
			>{\raggedright\arraybackslash}X
			ccccc
			>{\raggedright\arraybackslash}X@{}}
		\toprule
		\textbf{Model} & \textbf{Routing design} & \textbf{Depth} &
		\textbf{Layers} & \textbf{Routed} & \textbf{Top-\(\boldsymbol{k}\)} &
		\textbf{Shared} & \textbf{Evidence} \\
		\midrule
		OLMoE-1B-7B & routed-only & 16 & 4, 8, 16 & 64 & 8 & 0 &
		geometry; factorial; NLL \\
		Mixtral-8x7B-v0.1 & routed-only & 32 & 8, 16, 32 & 8 & 2 & 0
		& geometry; factorial; NLL \\
		DeepSeek-MoE-16B & shared-expert & 28 & 7, 14, 28 & 64 & 6 &
		2 & geometry; factorial; NLL \\
		Qwen3-30B-A3B-Base & routed-only & 48 & 12, 24, 41 &
		128 & 8 & 0 & geometry \\
		Gemma4-26B-A4B & shared-expert & 30 & 8, 15, 26 & 128 & 8 & 1
		& geometry \\
		Qwen3.6-35B-A3B & hybrid & 40 & 10, 20, 34 &
		256 & 8 & 1 & geometry \\
		\bottomrule
	\end{tabularx}
	\caption{Models and evidence coverage.}
	\label{tab:models}
\end{table*}

\subsection{The Factorial Design}\label{sec:factorial-design}

At route position~\(j\), we define four actors:
\begin{itemize}
\item \(S=S_{j-1}\): the actual preceding experts (the
  \textbf{context});
\item \(s=e_j\): the actual selected expert (the \textbf{candidate});
\item \(r\): the highest-scoring eligible expert outside the complete
  actual Top-\(k\) route (the \textbf{rival});
\item \(W_1,\ldots,W_M\): legal alternative contexts of the same
  length.
\end{itemize}

Crossing the two candidates with the two context choices yields the
four cells in Figure~\ref{fig:method}. They support three contrasts: candidate
quality, contextual opportunity, and their interaction.

\textbf{Candidate advantage.} The candidate advantage compares the
selected candidate with the rival under a fixed context:
\begin{equation}\label{eq:advantage}
\begin{aligned}
A_{\mathrm{actual}}
  &= q_x(s\mid S)-q_x(r\mid S),\\
A_{\mathrm{alt}}
  &= \tfrac{1}{M}{\textstyle\sum_m}
     \bigl[q_x(s\mid W_m)-q_x(r\mid W_m)\bigr].
\end{aligned}
\end{equation}
Positive \(A_{\mathrm{actual}}\) means the selected expert explains
more of the residual than the rival under the same context.

\textbf{Context effect.} The context effects measure how replacing the
actual context changes each fixed candidate's novelty:
\begin{equation}\label{eq:transplant}
\begin{aligned}
T_s
  &= q_x(s\mid S)-\tfrac{1}{M}{\textstyle\sum_m}q_x(s\mid W_m),\\
T_r
  &= q_x(r\mid S)-\tfrac{1}{M}{\textstyle\sum_m}q_x(r\mid W_m).
\end{aligned}
\end{equation}

\textbf{Interaction.} The candidate-by-context interaction is the
difference-in-differences
\begin{equation}\label{eq:interaction}
D=A_{\mathrm{actual}}-A_{\mathrm{alt}}=T_s-T_r.
\end{equation}
Positive~\(D\) means that the actual context increases the selected
candidate's advantage relative to the rival under the residual-novelty
metric; absolute amplification additionally requires \(T_s>0\).
Negative~\(D\) means that the actual co-selection context narrows this
advantage. Candidate quality itself is assessed by
\(A_{\mathrm{actual}}\).

\subsection{Models and Protocol}\label{sec:protocol}

We study six open MoEs spanning routed-only, shared-expert, and modern
high-expert-count designs (Table~\ref{tab:models}). The geometry survey contains 18 model--layer cells. The factorial
analysis contains 24 OLMoE/Mixtral cells; we apply the same protocol post
hoc to 15 DeepSeek shared-expert cells and report those results separately.
Frozen-route NLL interventions cover all 39 factorial cells.

Each position-resolved factorial cell uses up to 2{,}048 held-out tokens
per layer. Alternative contexts have the same length as the actual
context and follow fixed load-matching and exclusion rules; DeepSeek's
two shared experts remain fixed outside every routed set. Token metrics
are first averaged within source context, and 95\% intervals use 1{,}000
paired source-context bootstrap resamples to capture source-context
sampling variability for the analyzed checkpoints~\citep{efron1993bootstrap}.
The three-seed training study is summarized with mean \(\pm\) standard
deviation. Appendix~\ref{app:factorial-protocol} gives the full construction,
split, and weighting details; Appendices~\ref{app:factorial-results}--\ref{app:function}
report cell-level estimates and sensitivity analyses.

%% ================================================================
\section{Geometric Overlap and Route Coherence}\label{sec:overlap}
%% ================================================================

Geometric specialization contains two independent claims: expert
subspaces should separate globally, and actual routes should fit tokens
better than matched alternatives. We test them separately.

\subsection{Expert Subspaces Overlap}\label{sec:separation}

ESSI (Eq.~\ref{eq:essi}) remains near one for eligible experts in all
six models and 18 analyzed layers, ranging from 0.776 to 1.060 with
median 0.969 (Figure~\ref{fig:overlap}A). In centered directional
geometry, between-expert separation is therefore comparable to local
within-expert variation. This pattern is inconsistent with a hard
directional partition under the present metric; mean offsets and other
forms of specialization lie outside its scope. A supporting global-core
analysis gives the same picture: median expert overlap with a shared core
rises from roughly 0.27--0.44 at dimension 64 to 0.77--0.98 at dimension
512.

\subsection{Routes Remain Coherent}\label{sec:coherence}

Overlap need not make experts interchangeable. For each token,
we compare the normalized residual (Appendix~\ref{app:essi})
under the
actual route with that under a load-matched alternative of the same
size.

The alternative-to-actual ratio exceeds one in all 18 cells,
ranging from 1.012 to 1.901 with median 1.283
(Figure~\ref{fig:overlap}B). The router is therefore selective within
shared geometry: actual routes fit their tokens better even though their
experts do not occupy isolated regions.

Figure~\ref{fig:overlap} thus disfavors two limiting accounts: hard
directional partitions predict ESSI well above one, and
interchangeability predicts residual ratios near one; neither
prediction holds. The remaining pattern
is structured overlap, but route coherence still cannot distinguish
positive interaction from the selection of individually strong
candidates; Section~\ref{sec:complementarity} separates them.

\begin{figure*}[t]
	\centering
	\includegraphics[width=\textwidth]{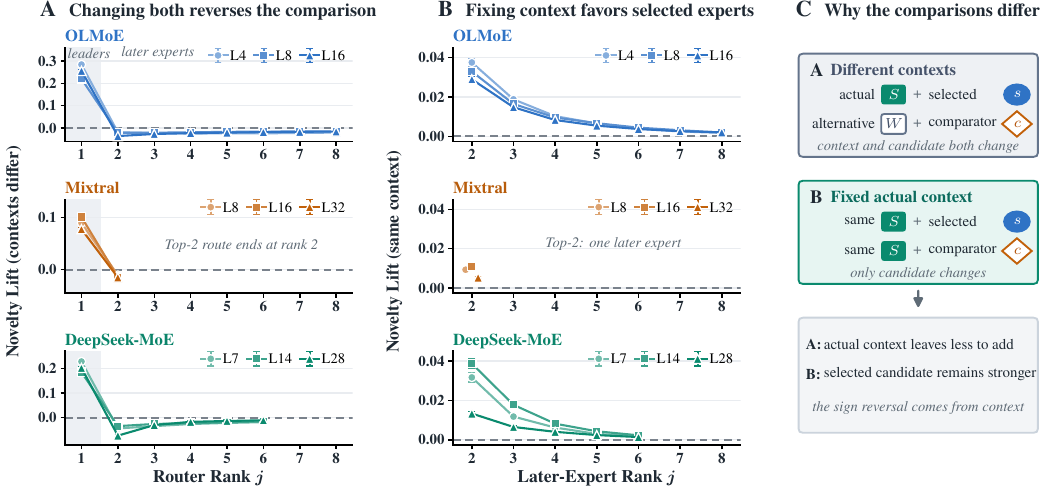}
	\caption{Fixing the context reverses the conclusion about
		later experts.
		A:~When the actual route is compared with five load-near
		control routes, both candidate and context change; all nine leader cells
		are positive but all 39~later-expert cells are negative.
		B:~When the actual preceding context is fixed, all 39~selected
		versus control-candidate cells favor the selected expert (mean lift
		0.001382--0.038642).
		C:~The two comparisons answer different questions: changing the
		context can make a strong candidate appear weak.
		DeepSeek's two shared experts remain fixed outside the routed context in
		both panels. Error bars are 95\% source-context bootstrap intervals.}
	\label{fig:reversal}
\end{figure*}

\begin{figure*}[t]
	\centering
	\includegraphics[width=\textwidth]{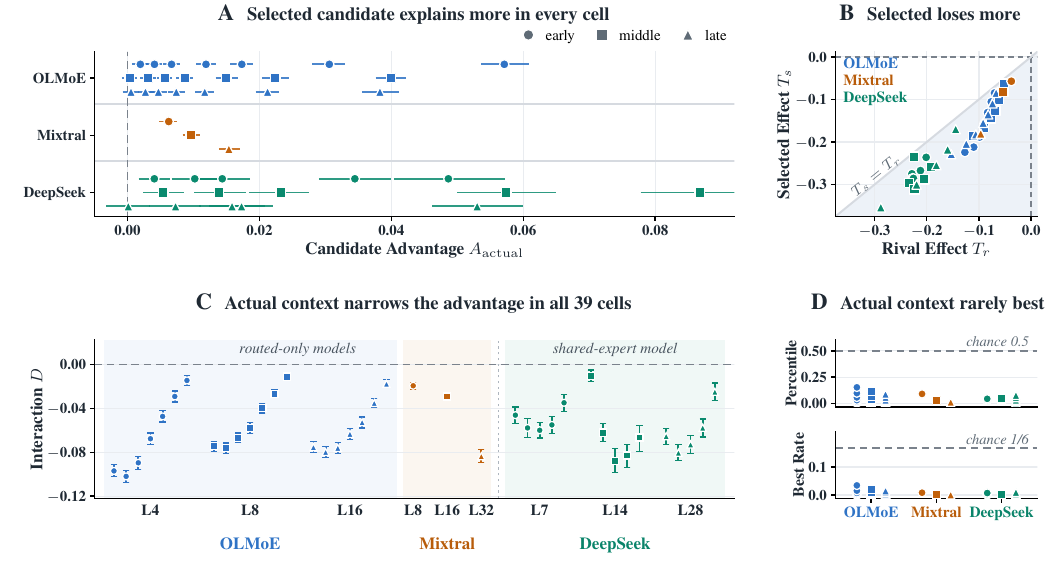}
	\caption{The selected candidate is strong, but its context
		narrows rather than amplifies its advantage.
		A:~Candidate advantage \(A_{\mathrm{actual}}>0\) in all
		39~point estimates. For OLMoE and Mixtral, 22 of 24~CIs exclude zero;
		DeepSeek's macro advantage is \(0.02616\) (95\% CI
		\([0.02412,0.02800]\)).
		B:~Both context effects are negative (\(T_s,T_r<0\)), and
		every cell lies below the \(T_s=T_r\) diagonal: the selected candidate
		loses more residual opportunity than the rival.
		C:~The interaction is negative in all 39 cells.
		D:~The selected candidate's novelty under the actual prefix
		ranks near the bottom across prefixes. Dashed lines show the chance
		center (0.5) and uniform best-rate (1/6). Error bars are 95\%
		source-context bootstrap intervals.}
	\label{fig:factorial}
\end{figure*}

%% ================================================================
\section{Strong Candidates, Negative Interaction}\label{sec:complementarity}
%% ================================================================

The factorial distinguishes complementary experts from merely the
strongest candidates for the token. We first show why changing candidate and context together can
reverse the apparent conclusion.

\subsection{Why Naive Comparisons Reverse}\label{sec:confounding}

In the changing-context diagnostic, we order each route by router score
and compare each selected expert's marginal geometric novelty with the
mean over five load-near, full-route-disjoint control routes, so both
the candidate and the preceding context change. These load-near control
candidates are distinct from the score-defined rival~\(r\) used in the
factorial below.

When both candidate and context change, all nine leader comparisons are
positive, whereas all 39 later-expert comparisons are negative
(Figure~\ref{fig:reversal}A). This comparison would therefore rank later
selected experts below the controls.

Because the comparison also changes residual opportunity, it cannot
isolate candidate quality. A coherent prefix may already cover what its
next candidate would add. Holding that prefix fixed and replacing only
the candidate reverses the result: all 39 comparisons favor the selected
expert, with mean lift from 0.001382 to 0.038642 and confidence intervals
above zero (Figure~\ref{fig:reversal}B).

The sign reversal shows that the two diagnostics answer different
questions (Figure~\ref{fig:reversal}C). The factorial next asks whether
the actual prefix increases the selected candidate's advantage relative
to alternative prefixes. Appendix~\ref{app:factorial-protocol} gives the control
construction.

\subsection{Factorial Results}\label{sec:routed-factorial}

Figure~\ref{fig:factorial} varies candidate and context independently
across 39 cells in OLMoE, Mixtral, and DeepSeek.

\textbf{The selected candidate explains more of the residual.} Under the actual
prefix, \(A_{\mathrm{actual}}>0\) in all 39 point estimates; 22 of 24
OLMoE/Mixtral intervals exclude zero, and DeepSeek's macro advantage is
\(0.02616\) (95\% CI \([0.02412,0.02800]\);
Figure~\ref{fig:factorial}A). With the prefix fixed, the selected expert explains more of the residual
representation than the highest-scoring eligible rival.
This establishes candidate quality; complementarity depends on the
interaction~\(D\).

\textbf{The actual prefix leaves less residual opportunity.} Every
\(T_s\) and \(T_r\) estimate is negative
(Figure~\ref{fig:factorial}B). For OLMoE/Mixtral, the ranges are
\(-0.2284\) to \(-0.0573\) for \(T_s\) and \(-0.1531\) to
\(-0.0379\) for \(T_r\); DeepSeek has \(T_s=-0.26864\) (95\% CI
\([-0.27433,-0.26271]\)) versus \(T_r=-0.21094\) (95\% CI
\([-0.21640,-0.20583]\)). Because each candidate is evaluated under prefixes of equal length,
these contrasts isolate the context's residual opportunity. The actual
prefix has already covered directions that either candidate could add.

\textbf{The actual prefix shrinks the selected candidate's advantage.}
The interaction is negative in all 39 cells, and every 95\% interval lies
below zero (Figure~\ref{fig:factorial}C). For OLMoE and Mixtral, \(D\)
ranges from \(-0.1020\) to \(-0.0111\), with a 24-cell macro estimate of
\(-0.05548\,[-0.05780,\,-0.05340]\). For DeepSeek, it ranges from
\(-0.08773\) to \(-0.01022\), with macro
\(D=-0.05770\) (95\% CI \([-0.06165,-0.05405]\)). Thus, the actual prefix removes more novelty from the selected candidate
than from the rival. The selected expert remains stronger, but by a
smaller margin under its actual context.

The selected candidate's actual-prefix novelty also ranks near the
bottom across alternatives (percentiles 0.006--0.154 for OLMoE/Mixtral
and 0.007--0.079 for DeepSeek; Figure~\ref{fig:factorial}D). This scale-free rank analysis is consistent with the interaction
estimates. Because route widths and expert pools differ,
magnitude comparisons are most meaningful within an architecture.
Descriptively, the interaction trends toward zero at later OLMoE
positions and becomes more negative across Mixtral's analyzed layers.

\subsection{Sensitivity Analyses}\label{sec:robustness}

We repeat the interaction analysis with raw geometric gain,
nearest-residual matching, and a strict fifth-percentile train-IQR
caliper; an independent CPU tall-SVD reconstruction provides a numerical
cross-check. Raw-gain and nearest-match intervals remain negative in all
24 OLMoE/Mixtral cells. Under the strict caliper, every point estimate
remains negative and 20 of 24 intervals exclude zero at 7.88\% coverage.
DeepSeek's macro interaction is also negative under all three variants:
\(-0.01966\,[-0.02096,\,-0.01846]\) under raw gain,
\(-0.06157\,[-0.06594,\,-0.05701]\) under nearest matching, and
\(-0.01780\,[-0.02472,\,-0.01099]\) under the strict caliper at
5.8--6.2\% per-layer coverage. The independent reconstruction agrees
within the specified tolerances (Appendix~\ref{app:robustness}).

Across these variants, the actual prefix consistently reduces the
selected candidate's relative linear novelty. Together with route
coherence and candidate quality, this pattern supports
\textbf{coherent overlap}: the router selects well-suited experts from a
shared token-conditioned geometric neighborhood. The next section asks
how this geometry relates to predictive value.

%% ================================================================
\section{Functional Value Within Overlapping Geometry}\label{sec:function}
%% ================================================================

\begin{figure*}[t]
    \centering
    \includegraphics[width=\textwidth]{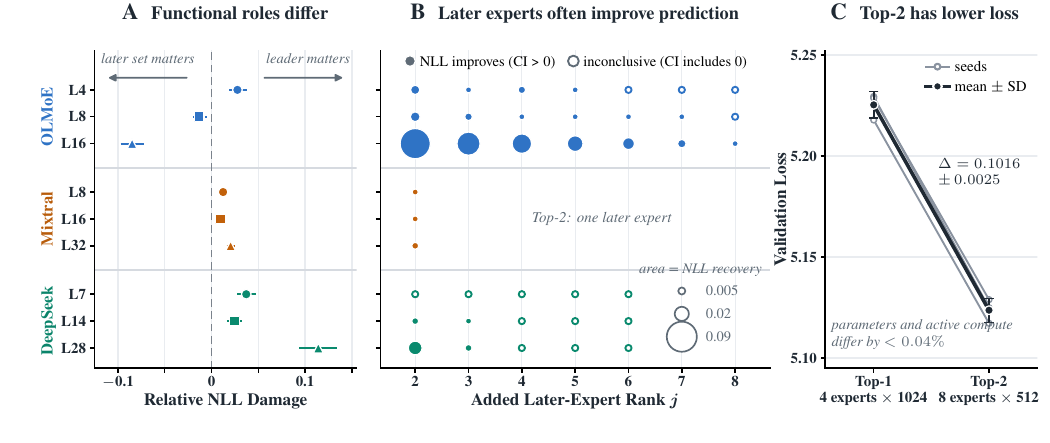}
    \caption{Later experts retain functional value within overlapping
    geometry.
    A:~Frozen-route replacement favors the leader in seven of
    nine cells, but the aggregate later set is more important at OLMoE
    layers~8 and~16.
    B:~Adding later experts reduces next-token NLL in 24 of 39~cases;
    the other 15 intervals overlap zero.
    C:~In the controlled Top-1/Top-2 study, Top-2 has lower validation
    loss in all three seeds (\(\Delta=0.1016\pm0.0025\), mean \(\pm\)
    SD).}
    \label{fig:function}
\end{figure*}

The factorial measures linear coverage of router inputs, whereas each
expert applies a parameterized nonlinear transformation. Similar input
subspaces can therefore support different features or output directions.
A negative geometric interaction characterizes residual linear coverage,
not the value of the resulting computation. We evaluate three functional
questions: whether later additions improve prediction, whether value is
concentrated in the leader, and whether a multi-expert advantage appears
under controlled compute.

\textbf{Many later-expert additions improve prediction.} The next
selected expert reduces next-token NLL in 24 of 39 frozen-route
additions; the remaining 15 intervals include zero
(Figure~\ref{fig:function}B). OLMoE is positive in 17 of 21 cases,
Mixtral is positive at all three layers, and DeepSeek has four positive
additions at early ranks. The largest effects are early: the first OLMoE
layer-16 addition recovers \(0.0906\,[0.0836,\,0.0975]\) NLL, while the
last recovers \(0.0030\,[0.0021,\,0.0039]\). Effect sizes decay with
router rank, and later estimates become smaller and less precise. The
pattern is declining marginal value rather than uniform redundancy: low
linear novelty can coexist with predictive gain (Appendix~\ref{app:function}).

\textbf{Functional value is distributed across the route.} Replacing the
leader causes more NLL damage than replacing the aggregate later set in seven of
nine configurations, confirming that router rank is informative. OLMoE
reverses at layers 8 and 16, where replacing the seven later experts
causes more aggregate damage than replacing the leader, by
\(0.0848\,[0.0724,\,0.0971]\) NLL at layer~16
(Figure~\ref{fig:function}A). This aggregate comparison shows that leader prominence can coexist with
substantial collective value among the later experts.

\textbf{Top-2 outperforms Top-1 under controlled compute.} In the controlled six-layer MoE, Top-1 (4 experts of width 1{,}024) and
Top-2 (8 experts of width 512) have the same active intermediate capacity;
total parameters and active compute differ by less than
0.04\%~\citep{clark2022scaling,ludziejewski2024finegrained}. Top-2
has lower validation loss in all three seeds, with
\(\Delta=0.1016\pm0.0025\) (Figure~\ref{fig:function}C). Because expert count and width change jointly, the comparison pertains
to this Top-2 configuration and controlled regime.

The probes are complementary: additions measure marginal value in a
pretrained route, replacement tests concentration in the leader, and
matched training permits adaptation. Geometry remains informative about route
organization and residual coverage, but it is incomplete as an importance
score because utility also depends on nonlinear computation within the
shared neighborhood.

%% ================================================================
\section{Potential Impact on MoE Analysis and Design}\label{sec:impact}
%% ================================================================

The framework separates three levels of evidence: a better route
establishes \emph{coherence}, a better candidate under a fixed prefix
establishes \emph{quality}, and only the interaction establishes
context-specific geometric synergy. Functional necessity still requires an
output-level counterfactual.

The negative interaction is consistent with geometric saturation rather
than route failure. Because the actual prefix is selected from the same
router input, it can remove directions available to both candidates. The
stronger selected candidate may therefore lose more marginal novelty while
remaining better under a fixed prefix. Changing both context and candidate
mixes quality with residual opportunity, explaining the reversal in
Figure~\ref{fig:reversal}.

Complementarity depends on both the representation used and the
operation considered. We measure
rank-128 linear coverage of router inputs before expert transformation;
overlap there can coexist with different nonlinear features, output
directions, or logit effects. Claims of overlap should therefore specify
their representation, rank, metric, and intervention. Semantic routing and
expert collaboration may reveal specialization that this subspace metric
does not resolve~\citep{olson2025semantic,tang2025collaboration}.

For compression, similarity can screen candidates, but pruning, merging, or
skipping should be tested under the retained route because removal changes
the context of other experts~\citep{lee2025stun,chen2025eac}. Separating
subspaces likewise need not improve prediction; adaptive-\(k\) routing could
instead estimate another expert's output gain subject to load and compute
constraints~\citep{zeng2024adamoe}.

Our conclusions concern a linear, rank-128 router-input metric, factorial
analyses of three architectures, and one small-scale matched-compute
setting. Appendix~\ref{app:provenance} provides additional scope and
interpretation details.

%% ================================================================
\section{Conclusion}\label{sec:conclusion}
%% ================================================================

Route quality alone does not reveal why several experts help. We make
complementarity testable: ESSI calibrates separation against
within-expert variation, the prefix-controlled \(2\times2\) factorial
identifies candidate, context, and interaction, and functional
interventions measure the predictive value of overlapping experts.

Across six MoEs, we find \textbf{coherent overlap}: eligible experts'
centered directional subspaces overlap under our metric and routes
remain token-coherent. Across all 39 factorial cells, selected candidates explain more of the
residual than their rivals, yet the interaction is negative. Actual context thus
narrows rather than amplifies geometric advantage, while frozen-route
interventions and a controlled training study show that later experts can
remain useful.

Separating expert overlap, route coherence, candidate quality,
contextual interaction, and functional value changes the mechanistic
picture: multi-expert benefit can arise from
distinct computations within a shared token-relevant neighborhood without
disjoint linear coverage.

\FloatBarrier
\clearpage
\bibliographystyle{plainnat}
\bibliography{references}

\FloatBarrier
\clearpage
\appendix
% Keep the appendix's S-numbering while avoiding duplicate
% hyperref destinations after counters are reset.
\setcounter{section}{0}
\setcounter{subsection}{0}
\setcounter{subsubsection}{0}
\setcounter{table}{0}
\setcounter{figure}{0}
\setcounter{equation}{0}
\renewcommand{\thesection}{S\arabic{section}}
\renewcommand{\thesubsection}{S\arabic{section}.\arabic{subsection}}
\renewcommand{\thesubsubsection}{S\arabic{section}.\arabic{subsection}.\arabic{subsubsection}}
\renewcommand{\thetable}{S\arabic{table}}
\renewcommand{\thefigure}{S\arabic{figure}}
\renewcommand{\theequation}{S\arabic{equation}}
\renewcommand{\theHsection}{app.\arabic{section}}
\renewcommand{\theHsubsection}{app.\arabic{section}.\arabic{subsection}}
\renewcommand{\theHsubsubsection}{app.\arabic{section}.\arabic{subsection}.\arabic{subsubsection}}
\renewcommand{\theHtable}{app.\arabic{table}}
\renewcommand{\theHfigure}{app.\arabic{figure}}
\renewcommand{\theHequation}{app.\arabic{equation}}

\section*{Appendix}
\vspace{-0.25em}
\suppressfloats[t]

\section{Scope, Notation, and Experimental Assets}\label{app:scope}

We reserve \emph{rival} for the highest-scoring eligible
unselected routed expert in the factorial experiment. Figure~\ref{fig:reversal} instead uses load-near \emph{control
candidates}; we keep the
two constructions distinct. Functional intervals that overlap zero are
called \emph{inconclusive}.
Unless a table header states otherwise, a scalar written as
\(\hat\theta\,[l,u]\) denotes a point estimate followed by its 95\%
confidence interval; mean \(\pm\) standard deviation is used for the
three-seed matched-training summary. In all tables, $L$ is the analyzed
transformer layer index, not the total model depth.

The pretrained-model geometry, factorial, and intervention analyses use
a fixed 8{,}192-record corpus containing 2{,}048 records each from C4,
CodeSearchNet-Python, the \texttt{ccdv/arxiv-summarization} collection,
and UltraChat 200k~\citep{raffel2020t5,husain2019codesearchnet,cohan2018discourse,
ding2023ultrachat}. The matched-compute study uses a separate
8{,}192-record corpus with the same general, code, and dialogue sources
and an ARC science component comprising 1{,}024 ARC-Easy and 1{,}024
ARC-Challenge training examples~\citep{clark2018arc}. Raw third-party
records are not redistributed.

\begin{table*}[t]
\centering
\begin{tabularx}{\textwidth}{@{}lccccccX@{}}
\toprule
Model & Analyzed layers & Routed & Top-$k$ & Shared & ESSI & Factorial & Coverage \\
\midrule
OLMoE-1B-7B-0125 & 4, 8, 16 & 64 & 8 & 0 & yes & $j=2{:}8$ & factorial and interventions \\
Mixtral-8x7B-v0.1 & 8, 16, 32 & 8 & 2 & 0 & yes & $j=2$ & factorial and interventions \\
DeepSeek-MoE-16B & 7, 14, 28 & 64 & 6 & 2 & yes & $j=2{:}6$ & factorial and interventions \\
Qwen3-30B-A3B-Base & 12, 24, 41 & 128 & 8 & 0 & yes & no & geometry survey \\
Gemma4-26B-A4B & 8, 15, 26 & 128 & 8 & 1 & yes & no & geometry survey \\
Qwen3.6-35B-A3B & 10, 20, 34 & 256 & 8 & 1 & yes & no & geometry survey \\
\bottomrule
\end{tabularx}
\caption{Model and analysis coverage.}
\label{tab:app-models}
\end{table*}

\section{ESSI and Route-Coherence Protocol}\label{app:essi}

\subsection{Centered global and local tangent subspaces}

For the ESSI survey, assignment weights are native Top-$k$ router
weights renormalized to sum to one within each token. An expert is
eligible once it has at least 2,048 selected tokens. Its global basis $G_e$ is
the leading rank-128 basis of a centered, routing-weighted PCA fit.
For each eligible expert, anchors and a candidate pool are sampled
without replacement in proportion to that expert's routing weights.
At anchor $a$, Euclidean distances are computed within the candidate
pool, the 256 nearest representations are centered, and a local
rank-128 PCA tangent basis $L_{e,a}$ is fitted. The local distance is
the normalized projection/chordal distance in Eq.~\ref{eq:grassmann}, a
standard Grassmannian metric~\citep{edelman1998geometry,ye2016schubert}.

The OLMoE, Mixtral, and DeepSeek analyses use 2,048 anchors and up to
8,192 candidate tokens per eligible expert. The Qwen3, Gemma4, and
Qwen3.6 analyses use 512 anchors and up to 4,096 candidate tokens.
The base random seed is zero, with the layer index added for
layer-specific sampling. The ESSI denominator is the routing-load-weighted
mean expert-level local tangent distance, with numerical floor $\epsilon=10^{-12}$.

\subsection{Route coherence}

For each token, route coherence compares residual energy under the
actual selected routed set with an equal-cardinality alternative set.
The residual uses this section's centered basis family, refit per run
with the same construction as $G_e$: for a routed set $E$ with
$U_E=\operatorname{orth}([G_e:e\in E])$ and $\mu_E$ the
route-weight-averaged mean of its experts' means (alternative sets
reuse the actual route's weights), the normalized residual of router
input $z$ is
$\|(I-U_EU_E^\top)(z-\mu_E)\|_2^2/\max(\|z-\mu_E\|_2^2,10^{-12})$. It is
distinct from the uncentered factorial quantity $\rho$ in
Eq.~\ref{eq:rho}. The OLMoE, Mixtral, and DeepSeek runs average token-level
residuals over up to 50{,}000 evaluation tokens with one load-matched
alternative set per token, restricting each set to its fitted experts;
the Qwen3, Gemma4, and Qwen3.6 runs lower expert eligibility to
1{,}024 support tokens, keep only tokens whose complete routed set is
fitted, draw five alternative sets per token, and average within source
context before the cell mean.
Table~\ref{tab:app-geometry} records the mean normalized residual for
each set and their ratio (columns Sel., Alt., and Alt./Sel.),
corresponding to panel~B of Figure~\ref{fig:overlap}. Ratios above one
mean that the alternative leaves more
unexplained router-input energy. Shared experts, where present, are held
fixed outside the routed-set comparison.

\begin{table*}[t]
\centering
\begin{tabular}{@{}lrrrrrrrr@{}}
\toprule
Model & $L$ & $|\mathcal E|$ & Between & Within & ESSI & Sel. & Alt. & Alt./Sel. \\
\midrule
OLMoE-1B-7B & 4 & 64 & 0.838 & 0.876 & 0.957 & 0.260 & 0.371 & 1.425 \\
OLMoE-1B-7B & 8 & 64 & 0.810 & 0.858 & 0.944 & 0.275 & 0.359 & 1.309 \\
OLMoE-1B-7B & 16 & 64 & 0.819 & 0.834 & 0.982 & 0.250 & 0.329 & 1.319 \\
Mixtral-8x7B & 8 & 8 & 0.700 & 0.894 & 0.783 & 0.736 & 0.796 & 1.082 \\
Mixtral-8x7B & 16 & 8 & 0.682 & 0.878 & 0.776 & 0.691 & 0.753 & 1.089 \\
Mixtral-8x7B & 32 & 8 & 0.734 & 0.853 & 0.861 & 0.525 & 0.567 & 1.081 \\
DeepSeek-MoE-16B & 7 & 61 & 0.781 & 0.826 & 0.946 & 0.102 & 0.193 & 1.901 \\
DeepSeek-MoE-16B & 14 & 62 & 0.780 & 0.819 & 0.951 & 0.078 & 0.142 & 1.824 \\
DeepSeek-MoE-16B & 28 & 47 & 0.797 & 0.790 & 1.008 & 0.085 & 0.149 & 1.757 \\
Qwen3-30B-A3B & 12 & 79 & 0.857 & 0.842 & 1.018 & 0.270 & 0.342 & 1.265 \\
Qwen3-30B-A3B & 24 & 82 & 0.850 & 0.836 & 1.016 & 0.289 & 0.350 & 1.211 \\
Qwen3-30B-A3B & 41 & 78 & 0.823 & 0.797 & 1.033 & 0.207 & 0.270 & 1.302 \\
Gemma4-26B-A4B & 8 & 90 & 0.674 & 0.715 & 0.942 & 0.009 & 0.011 & 1.124 \\
Gemma4-26B-A4B & 15 & 98 & 0.638 & 0.744 & 0.857 & 0.130 & 0.132 & 1.012 \\
Gemma4-26B-A4B & 26 & 83 & 0.820 & 0.803 & 1.021 & 0.233 & 0.283 & 1.216 \\
Qwen3.6-35B-A3B & 10 & 127 & 0.871 & 0.825 & 1.056 & 0.241 & 0.320 & 1.327 \\
Qwen3.6-35B-A3B & 20 & 131 & 0.875 & 0.825 & 1.060 & 0.278 & 0.342 & 1.232 \\
Qwen3.6-35B-A3B & 34 & 127 & 0.828 & 0.789 & 1.049 & 0.194 & 0.257 & 1.329 \\
\bottomrule
\end{tabular}
\caption{Complete ESSI and route-coherence results. $|\mathcal E|$ is
the number of eligible experts. Between is the
unweighted mean inter-expert normalized chordal distance; within is the
load-weighted local tangent dispersion. Sel. and alt. are normalized
residuals for actual and matched alternative routes.}
\label{tab:app-geometry}
\end{table*}

\subsection{Shared-core overlap}

As a complementary overlap diagnostic, let $U_0^{(c)}$ be the leading
$c$ directions of a centered PCA fit over all sampled router inputs in a
layer. For an eligible rank-$p$ expert basis $G_e$ ($p=128$), we compute
\begin{equation}
\operatorname{CoreOverlap}_c(e)=
\frac{1}{p}\bigl\|U_0^{(c)\top}G_e\bigr\|_F^2.
\end{equation}
This is the fraction of the expert subspace captured by the shared
layer-wide core. Table~\ref{tab:app-core-overlap} reports the median over
all eligible expert--layer bases in each model's three analyzed layers.
This diagnostic complements ESSI: it measures how much of each expert
basis lies in a common core, whereas ESSI compares between-expert
distance with local within-expert dispersion.

\begin{table*}[t]
\centering
\begin{tabular}{@{}lrrrr@{}}
\toprule
Model & $c=64$ & $c=128$ & $c=256$ & $c=512$ \\
\midrule
OLMoE-1B-7B & 0.320 & 0.501 & 0.687 & 0.851 \\
Mixtral-8x7B & 0.435 & 0.695 & 0.879 & 0.977 \\
DeepSeek-MoE-16B & 0.361 & 0.547 & 0.733 & 0.887 \\
Qwen3-30B-A3B & 0.292 & 0.474 & 0.661 & 0.818 \\
Gemma4-26B-A4B & 0.403 & 0.658 & 0.839 & 0.937 \\
Qwen3.6-35B-A3B & 0.269 & 0.422 & 0.594 & 0.766 \\
\bottomrule
\end{tabular}
\caption{Median expert overlap with a centered global core.}
\label{tab:app-core-overlap}
\end{table*}

\section{Factorial Residual-Attribution Protocol}\label{app:factorial-protocol}

\subsection{Uncentered bases for the factorial analysis}

The factorial experiment uses an uncentered rank-128
router-input basis $B_e$ for every routed expert with at least 1,024
train-split tokens, again using row-normalized native Top-$k$ assignment
weights. For a set $P$, the implementation forms the orthonormal union
$B(P)$ and divides residual energy by raw input energy. Rows with exactly
zero input energy are masked as invalid. Fractional novelty uses
$\max(\rho_x(P),10^{-12})$ as its numerical floor.

The split is by connected components of source-record contexts, with an
80/20 train/evaluation fraction and seed zero. Each layer uses up to
2,048 train calibration tokens and 2,048 held-out evaluation tokens,
with a minimum of 1,024 evaluation tokens. Candidate and rival are both
evaluated using the actual rank-$j$ router weight, so their comparison
changes expert identity rather than mixture scale.

\subsection{Load-near control diagnostic in Figure~3}

Figure~3 uses a load-near control diagnostic distinct from the rival-based
factorial.
For each held-out token, the code constructs complete alternative routes
that are disjoint from the actual route. At each position it orders
eligible experts by absolute routing-load difference from the selected
expert, keeps the nearest eight, and draws without replacement using the
fixed seed. DeepSeek's two shared experts remain fixed outside every
routed route.

The changing-context quantity in Figure~3A is
\begin{equation}
\delta_j^{\mathrm{ind}}=
q(s_j\mid S_{j-1})-
\frac{1}{5}\sum_{m=1}^{5}q(c_{m,j}\mid C_{m,j-1}),
\end{equation}
where each $C_m$ is a complete load-near, route-disjoint control route.
Thus both context and candidate change. The corresponding same-prefix quantity
in Figure~3B is
\begin{equation}
\delta_j^{\mathrm{same}}=
q(s_j\mid S_{j-1})-q(c_j\mid S_{j-1}),\qquad j\geq2,
\end{equation}
where $c_j$ is taken from the first complete route in a deterministic
32-route control pool. Both candidates are evaluated with the selected
expert's actual rank-$j$ weight. This $c_j$ is a load-near control candidate; the factorial analysis
uses the score-defined rival~$r$ instead.

\begin{table}[!ht]
\centering
\footnotesize
\setlength{\tabcolsep}{1.8pt}
\begin{tabular}{@{}ll>{\raggedright\arraybackslash}p{3.15cm}>{\raggedright\arraybackslash}p{3.15cm}>{\raggedright\arraybackslash}p{3.05cm}r@{}}
\toprule
Model & $L$ & \shortstack[l]{Independent leader\\95\% CI} & Changing-context later cells & Same-prefix cells & \shortstack{Later token-cells\\completed/requested} \\
\midrule
OLMoE & 4 & $0.2845\,[0.2760,\,0.2932]$ & 7/7 neg.; range $-0.0250$ to $-0.0193$ & 7/7 pos.; range $0.0018$ to $0.0375$ & 14,336/14,336 \\
OLMoE & 8 & $0.2203\,[0.2137,\,0.2268]$ & 7/7 neg.; range $-0.0215$ to $-0.0149$ & 7/7 pos.; range $0.0020$ to $0.0329$ & 14,336/14,336 \\
OLMoE & 16 & $0.2564\,[0.2489,\,0.2639]$ & 7/7 neg.; range $-0.0362$ to $-0.0159$ & 7/7 pos.; range $0.0021$ to $0.0290$ & 14,336/14,336 \\
Mixtral & 8 & $0.0897\,[0.0843,\,0.0958]$ & 1/1 neg.; $-0.0105$ & 1/1 pos.; $0.0093$ & 2,048/2,048 \\
Mixtral & 16 & $0.1011\,[0.0977,\,0.1047]$ & 1/1 neg.; $-0.0147$ & 1/1 pos.; $0.0109$ & 2,048/2,048 \\
Mixtral & 32 & $0.0777\,[0.0750,\,0.0807]$ & 1/1 neg.; $-0.0150$ & 1/1 pos.; $0.0054$ & 2,048/2,048 \\
DeepSeek & 7 & $0.2277\,[0.2189,\,0.2365]$ & 5/5 neg.; range $-0.0440$ to $-0.0185$ & 5/5 pos.; range $0.0020$ to $0.0316$ & 10,238/10,240 \\
DeepSeek & 14 & $0.1844\,[0.1775,\,0.1918]$ & 5/5 neg.; range $-0.0346$ to $-0.0151$ & 5/5 pos.; range $0.0023$ to $0.0386$ & 10,240/10,240 \\
DeepSeek & 28 & $0.2014\,[0.1950,\,0.2081]$ & 5/5 neg.; range $-0.0732$ to $-0.0108$ & 5/5 pos.; range $0.0014$ to $0.0133$ & 10,222/10,240 \\
\bottomrule
\end{tabular}
\caption{Complete model--layer summary for the Figure~3 confounding
 diagnostic. Later-cell and same-prefix entries report the number of
 rank-position cells with the indicated sign followed by the
 point-estimate range; completion counts refer to the later token-cells.}
\label{tab:app-confounding}
\end{table}

Across the changing-context diagnostic, all nine leader intervals are
above zero and all 39 later-expert intervals are below zero. Under the
fixed actual context, all 39 intervals are above zero, with point
estimates from 0.001382 to 0.038642 and aggregate coverage
79,852/79,872 token-cells. This sign reversal motivates the factorial: it shows
that candidate quality must be separated from the opportunity left by
its context.

\subsection{Rivals and alternative contexts}

The rival is fixed from the complete router-score ordering before any
context substitution: it is the highest-scoring eligible routed expert
outside the entire actual Top-$k$ route. OLMoE uses $M=5$ deterministic,
position-wise load-near, unique alternative full routes, whose
equal-length prefixes serve as the alternative contexts
$W_1,\ldots,W_M$ defined in Section~\ref{sec:factorial-design}. They exclude
the complete actual route and the rival, use a neighbor window of eight,
and reuse the same full-route prefixes across $j=2,\ldots,8$. Mixtral
has only one preceding expert at $j=2$ and uses all five eligible
alternative singleton leaders. DeepSeek uses the OLMoE construction for
$j=2,\ldots,6$; its two always-on shared experts stay fixed and never
enter routed IDs, candidate pools, rivals, alternative routes, or fit
codebooks, and never count toward prefix lengths.

Token-level quantities are averaged within source context. Confidence
intervals use 1,000 paired bootstrap resamples over source contexts with
seed zero~\citep{efron1993bootstrap}. They quantify held-out context-sampling uncertainty for the analyzed
checkpoints. DeepSeek follows the same estimand, split rule, basis
construction, and statistics, with its shared experts fixed as described
above.

\section{Complete Factorial Estimates}\label{app:factorial-results}

Table~\ref{tab:app-factorial-all} gives every reported cell. $A_S$ is
the selected-minus-rival advantage under the actual context (written
$A_{\mathrm{actual}}$ in Section~\ref{sec:factorial-design}), $T_s$ and $T_r$
are the selected and rival context effects, and $D=T_s-T_r$ is the
candidate-by-context interaction. Percentile is the fraction of the five
alternative-context selected-candidate gains that are no greater than
the actual-context gain. Every interval is a paired 95\% source-context
bootstrap interval.

All 24 OLMoE/Mixtral $D$ intervals are below zero. The equal-cell
macro, computed with equal cell weights within each source context, is
$-0.05548\,[-0.05780,\,-0.05340]$. The OLMoE and Mixtral macros are
$-0.05712\,[-0.05957,\,-0.05476]$ and
$-0.04419\,[-0.04654,\,-0.04174]$, respectively. The 24 per-cell
intervals are nominal 95\% intervals without multiplicity correction.

All 15 DeepSeek intervals are below zero, with macro
$-0.05770\,[-0.06165,-0.05405]$. The point estimate of $A_S$ is positive
in all 39 cells. For DeepSeek, the macro candidate advantage is
$0.02616\,[0.02412,\,0.02800]$, and the macro context effects are
$T_s=-0.26864\,[-0.27433,\,-0.26271]$ and
$T_r=-0.21094\,[-0.21640,\,-0.20583]$.

\begin{landscape}
\pagestyle{landscapetable}
\footnotesize
\setlength{\tabcolsep}{2.0pt}
\begin{longtable}{@{}lllrllllr@{}}
\caption{Complete 39-cell factorial decomposition: 24 OLMoE/Mixtral cells and 15 DeepSeek cells.}
\label{tab:app-factorial-all}\\
\toprule
Model & $L$ & $j$ & Valid tokens & $A_S$ [95\% CI] & $T_s$ [95\% CI] & $T_r$ [95\% CI] & $D$ [95\% CI] & Pctl. \\
\midrule
\endfirsthead
\multicolumn{9}{l}{\tablename\ \thetable\ continued}\\
\toprule
Model & $L$ & $j$ & Valid tokens & $A_S$ [95\% CI] & $T_s$ [95\% CI] & $T_r$ [95\% CI] & $D$ [95\% CI] & Pctl. \\
\midrule
\endhead
\midrule
\multicolumn{9}{r}{Continued on next page}\\
\endfoot
\bottomrule
\endlastfoot
Mixtral & 8 & 2 & 2048 & $0.0063\,[0.0049,\,0.0075]$ & $-0.0573\,[-0.0633,\,-0.0518]$ & $-0.0379\,[-0.0421,\,-0.0342]$ & $-0.0194\,[-0.0226,\,-0.0163]$ & 0.092 \\
Mixtral & 16 & 2 & 2048 & $0.0096\,[0.0084,\,0.0109]$ & $-0.0823\,[-0.0861,\,-0.0785]$ & $-0.0532\,[-0.0556,\,-0.0509]$ & $-0.0291\,[-0.0319,\,-0.0263]$ & 0.028 \\
Mixtral & 32 & 2 & 2048 & $0.0154\,[0.0138,\,0.0170]$ & $-0.1801\,[-0.1867,\,-0.1740]$ & $-0.0965\,[-0.1004,\,-0.0929]$ & $-0.0835\,[-0.0892,\,-0.0776]$ & 0.013 \\
OLMoE & 4 & 2 & 2048 & $0.0571\,[0.0536,\,0.0608]$ & $-0.2239\,[-0.2327,\,-0.2146]$ & $-0.1269\,[-0.1347,\,-0.1193]$ & $-0.0970\,[-0.1023,\,-0.0914]$ & 0.006 \\
OLMoE & 4 & 3 & 2048 & $0.0306\,[0.0280,\,0.0330]$ & $-0.2120\,[-0.2217,\,-0.2028]$ & $-0.1101\,[-0.1179,\,-0.1023]$ & $-0.1020\,[-0.1078,\,-0.0964]$ & 0.014 \\
OLMoE & 4 & 4 & 2048 & $0.0173\,[0.0156,\,0.0191]$ & $-0.1883\,[-0.1970,\,-0.1792]$ & $-0.0987\,[-0.1059,\,-0.0914]$ & $-0.0896\,[-0.0954,\,-0.0839]$ & 0.019 \\
OLMoE & 4 & 5 & 2048 & $0.0119\,[0.0104,\,0.0134]$ & $-0.1578\,[-0.1665,\,-0.1490]$ & $-0.0901\,[-0.0976,\,-0.0831]$ & $-0.0677\,[-0.0729,\,-0.0625]$ & 0.033 \\
OLMoE & 4 & 6 & 2048 & $0.0067\,[0.0054,\,0.0079]$ & $-0.1307\,[-0.1395,\,-0.1221]$ & $-0.0834\,[-0.0909,\,-0.0763]$ & $-0.0473\,[-0.0523,\,-0.0423]$ & 0.058 \\
OLMoE & 4 & 7 & 2048 & $0.0041\,[0.0028,\,0.0053]$ & $-0.1055\,[-0.1135,\,-0.0977]$ & $-0.0764\,[-0.0829,\,-0.0699]$ & $-0.0292\,[-0.0343,\,-0.0242]$ & 0.099 \\
OLMoE & 4 & 8 & 2048 & $0.0020\,[0.0007,\,0.0031]$ & $-0.0848\,[-0.0921,\,-0.0771]$ & $-0.0702\,[-0.0770,\,-0.0641]$ & $-0.0146\,[-0.0194,\,-0.0100]$ & 0.154 \\
OLMoE & 8 & 2 & 2048 & $0.0399\,[0.0376,\,0.0422]$ & $-0.1856\,[-0.1930,\,-0.1782]$ & $-0.1112\,[-0.1169,\,-0.1055]$ & $-0.0743\,[-0.0790,\,-0.0706]$ & 0.011 \\
OLMoE & 8 & 3 & 2048 & $0.0224\,[0.0204,\,0.0245]$ & $-0.1667\,[-0.1739,\,-0.1594]$ & $-0.0905\,[-0.0960,\,-0.0858]$ & $-0.0762\,[-0.0810,\,-0.0716]$ & 0.011 \\
OLMoE & 8 & 4 & 2048 & $0.0150\,[0.0134,\,0.0168]$ & $-0.1442\,[-0.1510,\,-0.1376]$ & $-0.0773\,[-0.0821,\,-0.0729]$ & $-0.0669\,[-0.0714,\,-0.0625]$ & 0.017 \\
OLMoE & 8 & 5 & 2048 & $0.0087\,[0.0074,\,0.0099]$ & $-0.1262\,[-0.1326,\,-0.1192]$ & $-0.0684\,[-0.0727,\,-0.0642]$ & $-0.0577\,[-0.0627,\,-0.0531]$ & 0.025 \\
OLMoE & 8 & 6 & 2048 & $0.0056\,[0.0045,\,0.0067]$ & $-0.1013\,[-0.1069,\,-0.0955]$ & $-0.0614\,[-0.0652,\,-0.0572]$ & $-0.0399\,[-0.0442,\,-0.0353]$ & 0.043 \\
OLMoE & 8 & 7 & 2048 & $0.0031\,[0.0019,\,0.0043]$ & $-0.0829\,[-0.0879,\,-0.0777]$ & $-0.0562\,[-0.0601,\,-0.0524]$ & $-0.0266\,[-0.0304,\,-0.0230]$ & 0.067 \\
OLMoE & 8 & 8 & 2048 & $0.0003\,[-0.0009,\,0.0014]$ & $-0.0628\,[-0.0675,\,-0.0584]$ & $-0.0517\,[-0.0553,\,-0.0482]$ & $-0.0111\,[-0.0146,\,-0.0078]$ & 0.116 \\
OLMoE & 16 & 2 & 2048 & $0.0383\,[0.0355,\,0.0412]$ & $-0.2284\,[-0.2366,\,-0.2212]$ & $-0.1531\,[-0.1599,\,-0.1469]$ & $-0.0753\,[-0.0801,\,-0.0703]$ & 0.006 \\
OLMoE & 16 & 3 & 2048 & $0.0212\,[0.0194,\,0.0230]$ & $-0.2040\,[-0.2117,\,-0.1959]$ & $-0.1242\,[-0.1300,\,-0.1179]$ & $-0.0798\,[-0.0851,\,-0.0749]$ & 0.007 \\
OLMoE & 16 & 4 & 2048 & $0.0117\,[0.0103,\,0.0131]$ & $-0.1820\,[-0.1899,\,-0.1743]$ & $-0.1060\,[-0.1118,\,-0.1009]$ & $-0.0760\,[-0.0813,\,-0.0709]$ & 0.009 \\
OLMoE & 16 & 5 & 2048 & $0.0074\,[0.0061,\,0.0088]$ & $-0.1552\,[-0.1624,\,-0.1484]$ & $-0.0920\,[-0.0975,\,-0.0868]$ & $-0.0632\,[-0.0683,\,-0.0582]$ & 0.022 \\
OLMoE & 16 & 6 & 2048 & $0.0047\,[0.0034,\,0.0059]$ & $-0.1350\,[-0.1418,\,-0.1280]$ & $-0.0823\,[-0.0872,\,-0.0775]$ & $-0.0527\,[-0.0579,\,-0.0479]$ & 0.033 \\
OLMoE & 16 & 7 & 2048 & $0.0027\,[0.0015,\,0.0040]$ & $-0.1092\,[-0.1150,\,-0.1038]$ & $-0.0741\,[-0.0792,\,-0.0699]$ & $-0.0351\,[-0.0388,\,-0.0311]$ & 0.058 \\
OLMoE & 16 & 8 & 2048 & $0.0005\,[-0.0007,\,0.0018]$ & $-0.0847\,[-0.0899,\,-0.0794]$ & $-0.0671\,[-0.0716,\,-0.0629]$ & $-0.0176\,[-0.0213,\,-0.0138]$ & 0.089 \\
DeepSeek & 7 & 2 & 2048 & $0.0486\,[0.0404,\,0.0572]$ & $-0.2746\,[-0.2858,\,-0.2635]$ & $-0.2284\,[-0.2374,\,-0.2190]$ & $-0.0462\,[-0.0539,\,-0.0388]$ & 0.032 \\
DeepSeek & 7 & 3 & 2048 & $0.0345\,[0.0290,\,0.0399]$ & $-0.2849\,[-0.2973,\,-0.2735]$ & $-0.2271\,[-0.2373,\,-0.2169]$ & $-0.0578\,[-0.0666,\,-0.0493]$ & 0.026 \\
DeepSeek & 7 & 4 & 2048 & $0.0144\,[0.0104,\,0.0185]$ & $-0.2854\,[-0.2985,\,-0.2736]$ & $-0.2254\,[-0.2363,\,-0.2135]$ & $-0.0600\,[-0.0671,\,-0.0527]$ & 0.029 \\
DeepSeek & 7 & 5 & 2048 & $0.0102\,[0.0067,\,0.0136]$ & $-0.2671\,[-0.2790,\,-0.2551]$ & $-0.2120\,[-0.2224,\,-0.2013]$ & $-0.0551\,[-0.0627,\,-0.0473]$ & 0.035 \\
DeepSeek & 7 & 6 & 2048 & $0.0041\,[0.0017,\,0.0065]$ & $-0.2360\,[-0.2484,\,-0.2241]$ & $-0.2011\,[-0.2115,\,-0.1908]$ & $-0.0349\,[-0.0425,\,-0.0276]$ & 0.045 \\
DeepSeek & 14 & 2 & 2048 & $0.0868\,[0.0778,\,0.0959]$ & $-0.2344\,[-0.2417,\,-0.2268]$ & $-0.2241\,[-0.2323,\,-0.2159]$ & $-0.0102\,[-0.0156,\,-0.0051]$ & 0.026 \\
DeepSeek & 14 & 3 & 2048 & $0.0574\,[0.0500,\,0.0649]$ & $-0.2960\,[-0.3074,\,-0.2838]$ & $-0.2339\,[-0.2432,\,-0.2248]$ & $-0.0621\,[-0.0703,\,-0.0541]$ & 0.023 \\
DeepSeek & 14 & 4 & 2048 & $0.0232\,[0.0184,\,0.0275]$ & $-0.3111\,[-0.3228,\,-0.3000]$ & $-0.2234\,[-0.2329,\,-0.2137]$ & $-0.0877\,[-0.0984,\,-0.0768]$ & 0.021 \\
DeepSeek & 14 & 5 & 2048 & $0.0139\,[0.0101,\,0.0180]$ & $-0.2877\,[-0.2987,\,-0.2760]$ & $-0.2048\,[-0.2150,\,-0.1955]$ & $-0.0829\,[-0.0940,\,-0.0727]$ & 0.027 \\
DeepSeek & 14 & 6 & 2047 & $0.0054\,[0.0024,\,0.0085]$ & $-0.2584\,[-0.2710,\,-0.2471]$ & $-0.1919\,[-0.2013,\,-0.1827]$ & $-0.0666\,[-0.0790,\,-0.0554]$ & 0.044 \\
DeepSeek & 28 & 2 & 2048 & $0.0530\,[0.0462,\,0.0600]$ & $-0.3533\,[-0.3618,\,-0.3446]$ & $-0.2879\,[-0.2954,\,-0.2795]$ & $-0.0653\,[-0.0727,\,-0.0584]$ & 0.007 \\
DeepSeek & 28 & 3 & 2047 & $0.0173\,[0.0124,\,0.0221]$ & $-0.2997\,[-0.3084,\,-0.2905]$ & $-0.2194\,[-0.2283,\,-0.2106]$ & $-0.0803\,[-0.0877,\,-0.0730]$ & 0.018 \\
DeepSeek & 28 & 4 & 2046 & $0.0158\,[0.0108,\,0.0209]$ & $-0.2543\,[-0.2642,\,-0.2443]$ & $-0.1814\,[-0.1903,\,-0.1729]$ & $-0.0729\,[-0.0813,\,-0.0648]$ & 0.033 \\
DeepSeek & 28 & 5 & 2045 & $0.0073\,[0.0032,\,0.0111]$ & $-0.2175\,[-0.2271,\,-0.2069]$ & $-0.1600\,[-0.1692,\,-0.1517]$ & $-0.0575\,[-0.0660,\,-0.0497]$ & 0.056 \\
DeepSeek & 28 & 6 & 2043 & $0.0001\,[-0.0032,\,0.0034]$ & $-0.1694\,[-0.1782,\,-0.1604]$ & $-0.1445\,[-0.1532,\,-0.1361]$ & $-0.0249\,[-0.0330,\,-0.0170]$ & 0.079 \\
\end{longtable}
\normalsize

\section{Sensitivity Analyses and Numerical Validation}\label{app:robustness}

The sensitivity analyses change the scale or opportunity control while
holding candidate identity and context pairing fixed. Raw gain removes the
fractional normalization. Nearest matching pairs actual and alternative
contexts by train-IQR-scaled residual opportunity. The strict caliper
threshold is the fifth percentile of training-set nearest-match distances
after IQR scaling.
Table~\ref{tab:app-robust-all} reports all cell estimates.
\footnotesize
\setlength{\tabcolsep}{2.5pt}
\begin{longtable}{@{}lllrrrr@{}}
\caption{Cell-level interaction sensitivity analyses. All entries are estimates with paired 95\% source-context bootstrap intervals.}
\label{tab:app-robust-all}\\
\toprule
Model & $L$ & $j$ & Fractional novelty & Raw gain & Nearest residual & 5th-pctl. caliper \\
\midrule
\endfirsthead
\multicolumn{7}{l}{\tablename\ \thetable\ continued}\\
\toprule
Model & $L$ & $j$ & Fractional novelty & Raw gain & Nearest residual & 5th-pctl. caliper \\
\midrule
\endhead
\midrule
\multicolumn{7}{r}{Continued on next page}\\
\endfoot
\bottomrule
\endlastfoot
Mixtral & 8 & 2 & $-0.0194\,[-0.0226,\,-0.0163]$ & $-0.0152\,[-0.0176,\,-0.0131]$ & $-0.0152\,[-0.0178,\,-0.0128]$ & $-0.0010\,[-0.0032,\,0.0011]$ \\
Mixtral & 16 & 2 & $-0.0291\,[-0.0319,\,-0.0263]$ & $-0.0221\,[-0.0242,\,-0.0200]$ & $-0.0256\,[-0.0285,\,-0.0226]$ & $-0.0040\,[-0.0070,\,-0.0009]$ \\
Mixtral & 32 & 2 & $-0.0835\,[-0.0892,\,-0.0776]$ & $-0.0297\,[-0.0322,\,-0.0273]$ & $-0.0791\,[-0.0853,\,-0.0733]$ & $-0.0066\,[-0.0117,\,-0.0010]$ \\
OLMoE & 4 & 2 & $-0.0970\,[-0.1023,\,-0.0914]$ & $-0.0936\,[-0.0985,\,-0.0889]$ & $-0.0942\,[-0.0995,\,-0.0891]$ & $-0.0119\,[-0.0191,\,-0.0057]$ \\
OLMoE & 4 & 3 & $-0.1020\,[-0.1078,\,-0.0964]$ & $-0.0784\,[-0.0826,\,-0.0742]$ & $-0.0942\,[-0.0996,\,-0.0894]$ & $-0.0172\,[-0.0238,\,-0.0113]$ \\
OLMoE & 4 & 4 & $-0.0896\,[-0.0954,\,-0.0839]$ & $-0.0589\,[-0.0626,\,-0.0553]$ & $-0.0796\,[-0.0853,\,-0.0742]$ & $-0.0132\,[-0.0206,\,-0.0057]$ \\
OLMoE & 4 & 5 & $-0.0677\,[-0.0729,\,-0.0625]$ & $-0.0399\,[-0.0430,\,-0.0366]$ & $-0.0580\,[-0.0631,\,-0.0532]$ & $-0.0121\,[-0.0185,\,-0.0065]$ \\
OLMoE & 4 & 6 & $-0.0473\,[-0.0523,\,-0.0423]$ & $-0.0243\,[-0.0269,\,-0.0217]$ & $-0.0408\,[-0.0456,\,-0.0366]$ & $-0.0138\,[-0.0220,\,-0.0061]$ \\
OLMoE & 4 & 7 & $-0.0292\,[-0.0343,\,-0.0242]$ & $-0.0132\,[-0.0154,\,-0.0109]$ & $-0.0240\,[-0.0284,\,-0.0196]$ & $-0.0066\,[-0.0150,\,0.0019]$ \\
OLMoE & 4 & 8 & $-0.0146\,[-0.0194,\,-0.0100]$ & $-0.0061\,[-0.0079,\,-0.0041]$ & $-0.0137\,[-0.0184,\,-0.0098]$ & $-0.0055\,[-0.0129,\,0.0018]$ \\
OLMoE & 8 & 2 & $-0.0743\,[-0.0790,\,-0.0706]$ & $-0.0669\,[-0.0703,\,-0.0634]$ & $-0.0713\,[-0.0760,\,-0.0668]$ & $-0.0133\,[-0.0177,\,-0.0089]$ \\
OLMoE & 8 & 3 & $-0.0762\,[-0.0810,\,-0.0716]$ & $-0.0557\,[-0.0589,\,-0.0525]$ & $-0.0697\,[-0.0744,\,-0.0650]$ & $-0.0129\,[-0.0185,\,-0.0074]$ \\
OLMoE & 8 & 4 & $-0.0669\,[-0.0714,\,-0.0625]$ & $-0.0423\,[-0.0450,\,-0.0394]$ & $-0.0603\,[-0.0652,\,-0.0559]$ & $-0.0087\,[-0.0146,\,-0.0029]$ \\
OLMoE & 8 & 5 & $-0.0577\,[-0.0627,\,-0.0531]$ & $-0.0316\,[-0.0340,\,-0.0292]$ & $-0.0508\,[-0.0556,\,-0.0462]$ & $-0.0251\,[-0.0450,\,-0.0111]$ \\
OLMoE & 8 & 6 & $-0.0399\,[-0.0442,\,-0.0353]$ & $-0.0193\,[-0.0216,\,-0.0174]$ & $-0.0348\,[-0.0398,\,-0.0300]$ & $-0.0230\,[-0.0439,\,-0.0083]$ \\
OLMoE & 8 & 7 & $-0.0266\,[-0.0304,\,-0.0230]$ & $-0.0112\,[-0.0129,\,-0.0097]$ & $-0.0234\,[-0.0274,\,-0.0194]$ & $-0.0168\,[-0.0314,\,-0.0048]$ \\
OLMoE & 8 & 8 & $-0.0111\,[-0.0146,\,-0.0078]$ & $-0.0040\,[-0.0053,\,-0.0028]$ & $-0.0104\,[-0.0140,\,-0.0070]$ & $-0.0091\,[-0.0190,\,-0.0007]$ \\
OLMoE & 16 & 2 & $-0.0753\,[-0.0801,\,-0.0703]$ & $-0.0662\,[-0.0699,\,-0.0624]$ & $-0.0716\,[-0.0765,\,-0.0672]$ & $-0.0181\,[-0.0247,\,-0.0114]$ \\
OLMoE & 16 & 3 & $-0.0798\,[-0.0851,\,-0.0749]$ & $-0.0554\,[-0.0587,\,-0.0524]$ & $-0.0742\,[-0.0787,\,-0.0694]$ & $-0.0221\,[-0.0281,\,-0.0164]$ \\
OLMoE & 16 & 4 & $-0.0760\,[-0.0813,\,-0.0709]$ & $-0.0450\,[-0.0482,\,-0.0423]$ & $-0.0694\,[-0.0745,\,-0.0644]$ & $-0.0211\,[-0.0270,\,-0.0161]$ \\
OLMoE & 16 & 5 & $-0.0632\,[-0.0683,\,-0.0582]$ & $-0.0326\,[-0.0353,\,-0.0300]$ & $-0.0562\,[-0.0608,\,-0.0519]$ & $-0.0156\,[-0.0220,\,-0.0099]$ \\
OLMoE & 16 & 6 & $-0.0527\,[-0.0579,\,-0.0479]$ & $-0.0240\,[-0.0261,\,-0.0218]$ & $-0.0464\,[-0.0516,\,-0.0420]$ & $-0.0099\,[-0.0158,\,-0.0042]$ \\
OLMoE & 16 & 7 & $-0.0351\,[-0.0388,\,-0.0311]$ & $-0.0145\,[-0.0161,\,-0.0129]$ & $-0.0304\,[-0.0344,\,-0.0269]$ & $-0.0060\,[-0.0116,\,-0.0001]$ \\
OLMoE & 16 & 8 & $-0.0176\,[-0.0213,\,-0.0138]$ & $-0.0063\,[-0.0076,\,-0.0050]$ & $-0.0145\,[-0.0178,\,-0.0113]$ & $-0.0035\,[-0.0091,\,0.0021]$ \\
DeepSeek & 7 & 2 & $-0.0462\,[-0.0539,\,-0.0388]$ & $-0.0259\,[-0.0292,\,-0.0225]$ & $-0.0522\,[-0.0607,\,-0.0440]$ & $-0.0005\,[-0.0071,\,0.0068]$ \\
DeepSeek & 7 & 3 & $-0.0578\,[-0.0666,\,-0.0493]$ & $-0.0250\,[-0.0283,\,-0.0217]$ & $-0.0616\,[-0.0698,\,-0.0527]$ & $-0.0045\,[-0.0157,\,0.0059]$ \\
DeepSeek & 7 & 4 & $-0.0600\,[-0.0671,\,-0.0527]$ & $-0.0176\,[-0.0203,\,-0.0153]$ & $-0.0615\,[-0.0696,\,-0.0541]$ & $-0.0383\,[-0.0623,\,-0.0164]$ \\
DeepSeek & 7 & 5 & $-0.0551\,[-0.0627,\,-0.0473]$ & $-0.0130\,[-0.0150,\,-0.0109]$ & $-0.0596\,[-0.0678,\,-0.0517]$ & $-0.1091\,[-0.1507,\,-0.0677]$ \\
DeepSeek & 7 & 6 & $-0.0349\,[-0.0425,\,-0.0276]$ & $-0.0057\,[-0.0074,\,-0.0040]$ & $-0.0409\,[-0.0497,\,-0.0327]$ & $-0.0948\,[-0.1342,\,-0.0516]$ \\
DeepSeek & 14 & 2 & $-0.0102\,[-0.0156,\,-0.0051]$ & $-0.0291\,[-0.0331,\,-0.0255]$ & $-0.0231\,[-0.0294,\,-0.0172]$ & $-0.0018\,[-0.0087,\,0.0054]$ \\
DeepSeek & 14 & 3 & $-0.0621\,[-0.0703,\,-0.0541]$ & $-0.0272\,[-0.0305,\,-0.0242]$ & $-0.0750\,[-0.0868,\,-0.0633]$ & $-0.0133\,[-0.0296,\,-0.0005]$ \\
DeepSeek & 14 & 4 & $-0.0877\,[-0.0984,\,-0.0768]$ & $-0.0227\,[-0.0255,\,-0.0200]$ & $-0.0969\,[-0.1102,\,-0.0861]$ & $-0.0356\,[-0.0596,\,-0.0148]$ \\
DeepSeek & 14 & 5 & $-0.0829\,[-0.0940,\,-0.0727]$ & $-0.0159\,[-0.0179,\,-0.0140]$ & $-0.0861\,[-0.0970,\,-0.0750]$ & $-0.0317\,[-0.0509,\,-0.0151]$ \\
DeepSeek & 14 & 6 & $-0.0666\,[-0.0790,\,-0.0554]$ & $-0.0094\,[-0.0111,\,-0.0077]$ & $-0.0682\,[-0.0808,\,-0.0570]$ & $-0.0092\,[-0.0252,\,0.0044]$ \\
DeepSeek & 28 & 2 & $-0.0653\,[-0.0727,\,-0.0584]$ & $-0.0371\,[-0.0409,\,-0.0335]$ & $-0.0774\,[-0.0845,\,-0.0701]$ & $-0.0094\,[-0.0189,\,-0.0007]$ \\
DeepSeek & 28 & 3 & $-0.0803\,[-0.0877,\,-0.0730]$ & $-0.0271\,[-0.0300,\,-0.0245]$ & $-0.0773\,[-0.0850,\,-0.0703]$ & $-0.0130\,[-0.0223,\,-0.0047]$ \\
DeepSeek & 28 & 4 & $-0.0729\,[-0.0813,\,-0.0648]$ & $-0.0215\,[-0.0242,\,-0.0188]$ & $-0.0673\,[-0.0765,\,-0.0593]$ & $0.0008\,[-0.0082,\,0.0098]$ \\
DeepSeek & 28 & 5 & $-0.0575\,[-0.0660,\,-0.0497]$ & $-0.0133\,[-0.0153,\,-0.0113]$ & $-0.0531\,[-0.0612,\,-0.0457]$ & $-0.0034\,[-0.0138,\,0.0063]$ \\
DeepSeek & 28 & 6 & $-0.0249\,[-0.0330,\,-0.0170]$ & $-0.0041\,[-0.0055,\,-0.0027]$ & $-0.0224\,[-0.0296,\,-0.0153]$ & $-0.0026\,[-0.0121,\,0.0082]$ \\
\end{longtable}
\normalsize
\end{landscape}
\pagestyle{fancy}

For the 24 OLMoE/Mixtral cells, raw-gain and nearest-residual intervals
are negative in every cell. The strict-caliper point estimate is also
negative in every cell, with 20 of 24 intervals below zero. The caliper
retains 3,875 of 49,152 token-cells (7.88\%), so these estimates serve as
a low-coverage sensitivity analysis. DeepSeek's layer-level caliper coverages are
5.8\%, 5.9\%, and 6.2\% at layers 7, 14, and 28. Its macro estimates are
$-0.01966\,[-0.02096,-0.01846]$ for raw gain,
$-0.06157\,[-0.06594,-0.05701]$ for nearest matching, and
$-0.01780\,[-0.02472,-0.01099]$ for the strict caliper.

\begin{table}[!ht]
\centering
\begin{tabular}{@{}llrrrr@{}}
\toprule
Model & $L$ & Max residual & Max projector & Rank mism. & Nonfinite \\
\midrule
DeepSeek & 7 & $9.93\times10^{-8}$ & $1.00\times10^{-4}$ & 0 & 0 \\
DeepSeek & 14 & $8.19\times10^{-8}$ & $2.14\times10^{-4}$ & 0 & 0 \\
DeepSeek & 28 & $1.27\times10^{-7}$ & $1.27\times10^{-4}$ & 0 & 0 \\
Mixtral & 8 & $2.10\times10^{-7}$ & $1.18\times10^{-5}$ & 0 & 0 \\
Mixtral & 16 & $1.53\times10^{-7}$ & $2.36\times10^{-5}$ & 0 & 0 \\
Mixtral & 32 & $7.79\times10^{-8}$ & $1.02\times10^{-4}$ & 0 & 0 \\
OLMoE & 4 & $1.25\times10^{-7}$ & $1.77\times10^{-4}$ & 0 & 0 \\
OLMoE & 8 & $1.55\times10^{-7}$ & $1.12\times10^{-4}$ & 0 & 0 \\
OLMoE & 16 & $1.83\times10^{-7}$ & $9.78\times10^{-5}$ & 0 & 0 \\
\bottomrule
\end{tabular}
\caption{Independent CPU tall-SVD numerical validation. Tolerances are
$2\times10^{-6}$ for residual discrepancy and $5\times10^{-4}$ for
projector discrepancy, with zero allowed rank mismatches and nonfinite
comparisons.}
\label{tab:app-numerical}
\end{table}

Across the nine model--layer runs, the largest OLMoE/Mixtral residual
and projector discrepancies are $2.10\times10^{-7}$ and
$1.77\times10^{-4}$; the largest DeepSeek values are
$1.27\times10^{-7}$ and $2.14\times10^{-4}$. These values remain below
the specified tolerances, with zero rank mismatches and zero nonfinite
comparisons.

\FloatBarrier
\section{Functional Interventions and Matched Training}\label{app:function}

\subsection{Leader-versus-later replacement}

Positive values in Table~\ref{tab:app-replacement} mean replacing the
leader increases NLL more than replacing all later experts together;
negative values mean replacing all later experts together produces the
larger NLL increase.

\begin{table}[!ht]
\centering
\begin{tabular}{@{}llrl@{}}
\toprule
Model & $L$ & Difference [95\% CI] & Interpretation \\
\midrule
DeepSeek & 7 & $0.03712\,[0.02729,\,0.04730]$ & Leader larger \\
DeepSeek & 14 & $0.02475\,[0.01652,\,0.03312]$ & Leader larger \\
DeepSeek & 28 & $0.11410\,[0.09389,\,0.13382]$ & Leader larger \\
Mixtral & 8 & $0.01239\,[0.00804,\,0.01741]$ & Leader larger \\
Mixtral & 16 & $0.00934\,[0.00626,\,0.01244]$ & Leader larger \\
Mixtral & 32 & $0.02041\,[0.01626,\,0.02481]$ & Leader larger \\
OLMoE & 4 & $0.02769\,[0.01917,\,0.03796]$ & Leader larger \\
OLMoE & 8 & $-0.01323\,[-0.02000,\,-0.00472]$ & Later set larger \\
OLMoE & 16 & $-0.08483\,[-0.09712,\,-0.07241]$ & Later set larger \\
\bottomrule
\end{tabular}
\caption{Leader replacement damage minus aggregate later-expert
replacement damage.}
\label{tab:app-replacement}
\end{table}

\subsection{Adjacent ordered-prefix NLL recovery}

Table~\ref{tab:app-nll-additions} reports the NLL recovery from adding
the next selected expert in router-score order while freezing the route.
We classify an estimate as \emph{positive} when its 95\% interval is
above zero and as \emph{inconclusive} when the interval overlaps zero.
Across the 39 additions, 24 are positive and 15 are inconclusive.

\begingroup
\small
\setlength{\tabcolsep}{4pt}
\begin{longtable}{@{}llcll@{}}
\caption{Complete adjacent ordered-prefix NLL recovery results.}
\label{tab:app-nll-additions}\\
\toprule
Model & $L$ & Added rank & NLL recovery [95\% CI] & Classification \\
\midrule
\endfirsthead
\multicolumn{5}{l}{\tablename\ \thetable\ continued}\\
\toprule
Model & $L$ & Added rank & NLL recovery [95\% CI] & Classification \\
\midrule
\endhead
\midrule
\multicolumn{5}{r}{Continued on next page}\\
\endfoot
\bottomrule
\endlastfoot
DeepSeek & 7 & 1$\rightarrow$2 & $0.001122\,[-0.004068,\,0.006757]$ & Inconclusive \\
DeepSeek & 7 & 2$\rightarrow$3 & $-0.000018\,[-0.002618,\,0.002636]$ & Inconclusive \\
DeepSeek & 7 & 3$\rightarrow$4 & $0.000624\,[-0.000856,\,0.002333]$ & Inconclusive \\
DeepSeek & 7 & 4$\rightarrow$5 & $0.000093\,[-0.001081,\,0.001331]$ & Inconclusive \\
DeepSeek & 7 & 5$\rightarrow$6 & $0.000365\,[-0.000815,\,0.001487]$ & Inconclusive \\
DeepSeek & 14 & 1$\rightarrow$2 & $0.004953\,[0.001629,\,0.008037]$ & Positive \\
DeepSeek & 14 & 2$\rightarrow$3 & $0.002029\,[0.000235,\,0.003783]$ & Positive \\
DeepSeek & 14 & 3$\rightarrow$4 & $0.000261\,[-0.001062,\,0.001560]$ & Inconclusive \\
DeepSeek & 14 & 4$\rightarrow$5 & $0.000893\,[-0.000086,\,0.001735]$ & Inconclusive \\
DeepSeek & 14 & 5$\rightarrow$6 & $-0.000035\,[-0.000947,\,0.000811]$ & Inconclusive \\
DeepSeek & 28 & 1$\rightarrow$2 & $0.018939\,[0.015022,\,0.022850]$ & Positive \\
DeepSeek & 28 & 2$\rightarrow$3 & $0.005047\,[0.002449,\,0.007563]$ & Positive \\
DeepSeek & 28 & 3$\rightarrow$4 & $0.000654\,[-0.000996,\,0.002349]$ & Inconclusive \\
DeepSeek & 28 & 4$\rightarrow$5 & $-0.000027\,[-0.001178,\,0.001091]$ & Inconclusive \\
DeepSeek & 28 & 5$\rightarrow$6 & $0.000055\,[-0.000697,\,0.000791]$ & Inconclusive \\
Mixtral & 8 & 1$\rightarrow$2 & $0.003154\,[0.001639,\,0.004744]$ & Positive \\
Mixtral & 16 & 1$\rightarrow$2 & $0.002261\,[0.001238,\,0.003298]$ & Positive \\
Mixtral & 32 & 1$\rightarrow$2 & $0.004975\,[0.003630,\,0.006336]$ & Positive \\
OLMoE & 4 & 1$\rightarrow$2 & $0.008803\,[0.004683,\,0.012571]$ & Positive \\
OLMoE & 4 & 2$\rightarrow$3 & $0.003504\,[0.000941,\,0.005933]$ & Positive \\
OLMoE & 4 & 3$\rightarrow$4 & $0.005812\,[0.002789,\,0.009969]$ & Positive \\
OLMoE & 4 & 4$\rightarrow$5 & $0.002341\,[0.000839,\,0.004105]$ & Positive \\
OLMoE & 4 & 5$\rightarrow$6 & $0.000792\,[-0.000173,\,0.001787]$ & Inconclusive \\
OLMoE & 4 & 6$\rightarrow$7 & $0.001212\,[-0.000019,\,0.002631]$ & Inconclusive \\
OLMoE & 4 & 7$\rightarrow$8 & $-0.000827\,[-0.001812,\,0.000161]$ & Inconclusive \\
OLMoE & 8 & 1$\rightarrow$2 & $0.009869\,[0.005757,\,0.013230]$ & Positive \\
OLMoE & 8 & 2$\rightarrow$3 & $0.006029\,[0.004007,\,0.008225]$ & Positive \\
OLMoE & 8 & 3$\rightarrow$4 & $0.003793\,[0.002563,\,0.005010]$ & Positive \\
OLMoE & 8 & 4$\rightarrow$5 & $0.001859\,[0.000855,\,0.002859]$ & Positive \\
OLMoE & 8 & 5$\rightarrow$6 & $0.002158\,[0.001145,\,0.003192]$ & Positive \\
OLMoE & 8 & 6$\rightarrow$7 & $0.001280\,[0.000114,\,0.002893]$ & Positive \\
OLMoE & 8 & 7$\rightarrow$8 & $0.000130\,[-0.000686,\,0.001008]$ & Inconclusive \\
OLMoE & 16 & 1$\rightarrow$2 & $0.090607\,[0.083643,\,0.097480]$ & Positive \\
OLMoE & 16 & 2$\rightarrow$3 & $0.053769\,[0.049469,\,0.058214]$ & Positive \\
OLMoE & 16 & 3$\rightarrow$4 & $0.037903\,[0.034102,\,0.041830]$ & Positive \\
OLMoE & 16 & 4$\rightarrow$5 & $0.024718\,[0.022252,\,0.027490]$ & Positive \\
OLMoE & 16 & 5$\rightarrow$6 & $0.014026\,[0.011544,\,0.016559]$ & Positive \\
OLMoE & 16 & 6$\rightarrow$7 & $0.007353\,[0.006043,\,0.008780]$ & Positive \\
OLMoE & 16 & 7$\rightarrow$8 & $0.003032\,[0.002144,\,0.003919]$ & Positive \\
\end{longtable}
\endgroup

By model, OLMoE contributes 17 positive and four inconclusive
additions, Mixtral contributes three positive additions, and DeepSeek
contributes four positive and 11 inconclusive additions.

\subsection{Matched-compute Top-1 versus Top-2}

The controlled six-layer MoE comparison matches active intermediate
capacity exactly (1{,}024 units); total parameters (27{,}108{,}608
vs.\ 27{,}114{,}752) and recorded active FLOPs per token
(36{,}777{,}984 vs.\ 36{,}790{,}272) differ by less than 0.04\%. Top-1
uses four experts of width 1,024; Top-2 uses eight experts of width
512. Each seed trains for
20 million tokens (2,442 optimizer updates). Lower validation loss is
better.

\begin{table}[!ht]
\centering
\begin{tabular}{@{}rrrr@{}}
\toprule
Seed & Top-1 & Top-2 & Difference \\
\midrule
0 & 5.229414 & 5.124888 & 0.104527 \\ 
1 & 5.228661 & 5.128909 & 0.099751 \\ 
2 & 5.217812 & 5.117210 & 0.100602 \\ 
\midrule Mean $\pm$ SD & 5.225296 $\pm$ 0.006492 & 5.123669 $\pm$ 0.005944 & 0.101627 $\pm$ 0.002547 \\ 
\bottomrule
\end{tabular}
\caption{Matched-compute validation loss. Difference is Top-1 minus
Top-2, so positive values favor Top-2.}
\label{tab:app-matched}
\end{table}

Across the three seeds, Top-2 achieved lower validation loss, with a paired
difference of $0.101627\pm0.002547$ (mean $\pm$ sample SD).

\FloatBarrier
\section{Scope, Reproducibility, and Interpretation}\label{app:provenance}

\subsection{Scope of the evidence}

Our rank-128 linear geometry is defined on router-input representations.
Accordingly, \(D\) measures residual-geometric interaction; functional
consequences are evaluated separately through NLL interventions and
matched training. The complete factorial and functional analyses cover
OLMoE, Mixtral, and one shared-expert architecture, DeepSeek, while
Qwen3, Gemma4, and Qwen3.6 broaden the geometry survey. The
matched-training comparison controls active intermediate capacity,
parameters, and active FLOPs within the reported tolerances for the
tested six-layer model.

The score-ordered prefix is an analytical device, because standard
Top-\(k\) experts execute in parallel. Alternative prefixes satisfy the
matching and exclusion constraints, and the rival is the highest-scoring
eligible routed expert outside the actual route. These constructions
isolate candidate-by-context interaction.

\subsection{Reproducibility}

The factorial analyses use seed zero, source-record connected-component
splits, train-only expert fits, and context-level paired bootstrap
intervals. Shared experts remain fixed in every DeepSeek geometry and
functional comparison. Complete cell-level estimates and sensitivity
analyses are reported in Appendices~\ref{app:factorial-results}--\ref{app:function}.
The reported GPU experiments were conducted on a server equipped with eight
NVIDIA GeForce RTX 4090 GPUs (24{,}564 MiB per GPU). Each matched-compute
Top-1/Top-2 training run used a single GPU.

Centered ESSI and uncentered factorial bases answer different questions:
ESSI compares global expert separation with local tangent variation,
whereas \(D\) tests candidate-by-context interaction. NLL interventions
and matched training provide the complementary functional evidence.

\end{document}